%% file: acl2023.tex
\pdfoutput=1
\documentclass[11pt]{article}
\usepackage{ACL2023}
\usepackage{times}
\usepackage{latexsym}
\usepackage[T1]{fontenc}
\usepackage[utf8]{inputenc}
\usepackage{microtype}
\usepackage{inconsolata}
\usepackage{graphicx}
\usepackage{makecell}
\usepackage{hyperref}
\usepackage{microtype}
\usepackage{multirow}
\usepackage{booktabs}
\usepackage{siunitx}
\usepackage{amsmath}
\usepackage{algorithm}
\usepackage[noend]{algpseudocode}
\usepackage{amssymb}
\usepackage{bm}
\def\BState{\State\hskip-\ALG@thistlm}

\title{Topic-Guided Self-Introduction Generation for Social Media Users}

\author{Chunpu Xu\textsuperscript{\rm 1}, 
Jing Li\textsuperscript{\rm 1}\thanks{~~~Corresponding author}, 
Piji Li\textsuperscript{\rm 2}, Min Yang\textsuperscript{\rm 3}\\ 
\textsuperscript{\rm 1} Department of Computing, The Hong Kong Polytechnic University\\
\textsuperscript{\rm 2} College of Computer Science and Technology, \\
Nanjing University of Aeronautics and Astronautics\\
\textsuperscript{\rm 3} Shenzhen Key Laboratory for High Performance Data Mining, \\
Shenzhen Institutes of Advanced Technology, Chinese Academy of Sciences \\
\textsuperscript{\rm 1}\texttt{chun-pu.xu@connect.polyu.hk;}
\textsuperscript{\rm 1}\texttt{jing-amelia.li@polyu.edu.hk;}\\
\textsuperscript{\rm 2}\texttt{pjli@nuaa.edu.cn;}
\textsuperscript{\rm 3}\texttt{min.yang@siat.ac.cn}}

\begin{document}
\maketitle
\begin{abstract}
Millions of users are active on social media.
To allow users to better showcase themselves and network with others, we explore the auto-generation of social media \textit{self-introduction}, a short sentence outlining a user's personal interests. 
While most prior work profiles users with tags (e.g., ages), we investigate sentence-level self-introductions to provide a more natural and engaging way for users to know each other.  
Here we exploit a user's tweeting history to generate their self-introduction.
The task is non-trivial because the history content may be lengthy, noisy, and exhibit various personal interests.  
To address this challenge, we propose a novel unified topic-guided encoder-decoder (UTGED) framework; it models latent topics to reflect salient user interest, whose topic mixture then guides encoding a user's history and topic words control decoding their self-introduction.
For experiments, we collect a large-scale Twitter dataset, and extensive results show the superiority of our UTGED to the advanced encoder-decoder models without topic modeling.
\footnote{Our code and dataset are released at \url{https://github.com/cpaaax/UTGED}.}

\end{abstract}

\input{Sections/introduction.tex}
\input{Table/dataset_analysis.tex}
\input{Sections/Related_Work.tex}
\input{Sections/dataset_collection.tex}

\input{Sections/method.tex}

\input{Sections/Experimental_Setup.tex}

\input{Sections/Experimental_Results.tex}

\input{Sections/conclusion.tex}
\input{Sections/ack}

\input{Sections/limitation.tex}
\input{Sections/Ethics_Statement.tex}

\bibliography{anthology,custom}
\bibliographystyle{acl_natbib}

\newpage
\appendix
\input{Sections/appendix.tex}

\end{document}

%% file: Sections/introduction.tex
\section{Introduction}
The irresistible popularity of social media results in an explosive number of users, creating and broadcasting massive amounts of content every day.
Although it exhibits rich resources for users to build connections and share content, the sheer quantities of users might hinder one from finding those they want to follow
\cite{matikainen2015motivations}. 
To enable users to quickly know each other, many social platforms encourage a user to write a \textit{self-introduction}, a sentence to overview their personal interests.

A self-introduction is part of a self-described profile, which may else include locations, selfies, user tags, and so forth, and is crucial in online user interactions \cite{DBLP:books/daglib/p/McCay-PeetQ16}. 
Previous findings \cite{DBLP:conf/chi/HuttoYG13} indicate users tend to follow those displaying self-introductions 
because a well-written self-introduction will brief others about a user's interests 
and facilitate them to initialize connections.
It would benefit users in making like-minded friends and gaining popularity;
whereas not all users are skillful in writing a good self-introduction.
We are thus interested in how NLP may help and study \textbf{self-introduction generation}, a new application to learn user interests from their historical tweets (henceforth \textbf{user history}) and brief them in a  self-introduction.

\input{figure_sections/introduction_figure.tex}

Despite substantial efforts made in profiling users, most existing work \cite{DBLP:conf/acl/LiRH14, DBLP:conf/mir/FarseevNAC15, DBLP:conf/wsdm/FarnadiTCM18, DBLP:conf/ijcai/ChenGRHXGYZ19} focuses on \textit{extracting} keywords from user history and producing \textit{tag-level user attributes }
(e.g., interests, ages, and personality), which may later characterize personalization and recommendation
\cite{DBLP:conf/kdd/WangFXL19, DBLP:journals/tois/LiangLM22}. 
However, tag-level attributes profile a user through a fragmented view, while human readers may find it difficult to read.
On the contrary, we automate the writing of a \textit{sentence-level} self-introduction via \textit{language generation}, providing a more natural and easy-to-understand way to warm up social interactions.
It consequently will enable a better socializing experience and user engagement in social media. 

To practically train NLP models with capabilities in self-introduction writing, we collect a large-scale Twitter dataset with 170K public users. 
Each user presents a self-introduction (manually written by themselves) and previous tweets in their history, corresponding to a total of 10.2M tweets.

For methodology design, we take advantage of  cutting-edge practices using pre-trained encoder-decoder for language understanding and generation.
However, in real-world practice, users may post numerous tweets exhibiting \textit{lengthy content, noisy writings, and diverse interests}; these may challenge existing encoder-decoder models in capturing salient personal interests and reflecting them in the brief self-introduction writing.

To illustrate this challenge, Figure \ref{fig:intro_case} shows the self-introduction of a Twitter user $U$ and some sampled tweets from $U$'s user history.  
$U$ exhibits a mixture of interests varying in \textit{Delaware}, \textit{invertebrates}, \textit{paleontology}, \textit{museum}, and others, scatteredly indicated in multiple noisy tweets.
It presents a concrete challenge for models to digest the fragmented information, distill the introduction-worthy points, and condense them into a concise, coherent, and engaging self-introduction for further interactions.  
Moreover, existing NLP models are ineffective in encoding very long documents \cite{cao-wang-2022-hibrids}, whereas popular users may  post numerous tweets, resulting in a lengthy history to encode.

Consequently, we propose a novel unified topic-guided encoder-decoder (UTGED) framework for self-introduction generation. 
First, a neural topic model 
\cite{DBLP:conf/iclr/SrivastavaS17} clusters words by statistics to learn a mixture of latent topics in characterizing user  interests underlying their lengthy history.
Then, we inject the  latent topics into a BART-based encoder and decoder \cite{DBLP:conf/acl/LewisLGGMLSZ20}; 
the encoder employs topic distributions as continuous prompts \cite{DBLP:conf/emnlp/LesterAC21, DBLP:journals/corr/abs-2107-13586, DBLP:conf/acl/LiL20} to guide capturing personal interest mixture, 
and the decoder adopts topic words to control the writing for personalized self-introduction.



In experimental results, the comparison in both automatic and human evaluation   
show that UTGED  outperforms state-of-the-art encoder-decoder models without topic guidance; and ablation studies indicate the individual contribution from topic-guided encoder and decoder.
Then, we conduct parameter analyses on topic number and topic prompt length; they are followed by the study on model performance given users varying in historical tweet number, where UTGED consistently performs better. 
Finally, a case study and an error analysis interpret UTGED's superiority and limitations. 

\textit{To the best of our knowledge, we present the first NLP study on self-introduction writing from user tweeting history, where we build the first dataset for its empirical studies and show the benefits from latent topics to the state-of-the-art encoder-decoder paradigm.}
Below are details of our contributions.

$\bullet$ We present a new application to capture personal interests from a user's tweeting history and generate their self-introductions accordingly.

$\bullet$ We approach the application with a novel UTGED (unified topic-guided encoder-decoder) framework, which explores latent topics to represent users' personal interests and to jointly guide user encoding and self-introduction decoding. 

$\bullet$ 
We construct a large-scale Twitter dataset for self-introduction study and extensive experimental results on it show the superiority of UTGED practically and the benefits of latent topics on the task.

%% file: figure_sections/introduction_figure.tex
\begin{figure}[t]
\centering
\includegraphics[width=1\columnwidth]{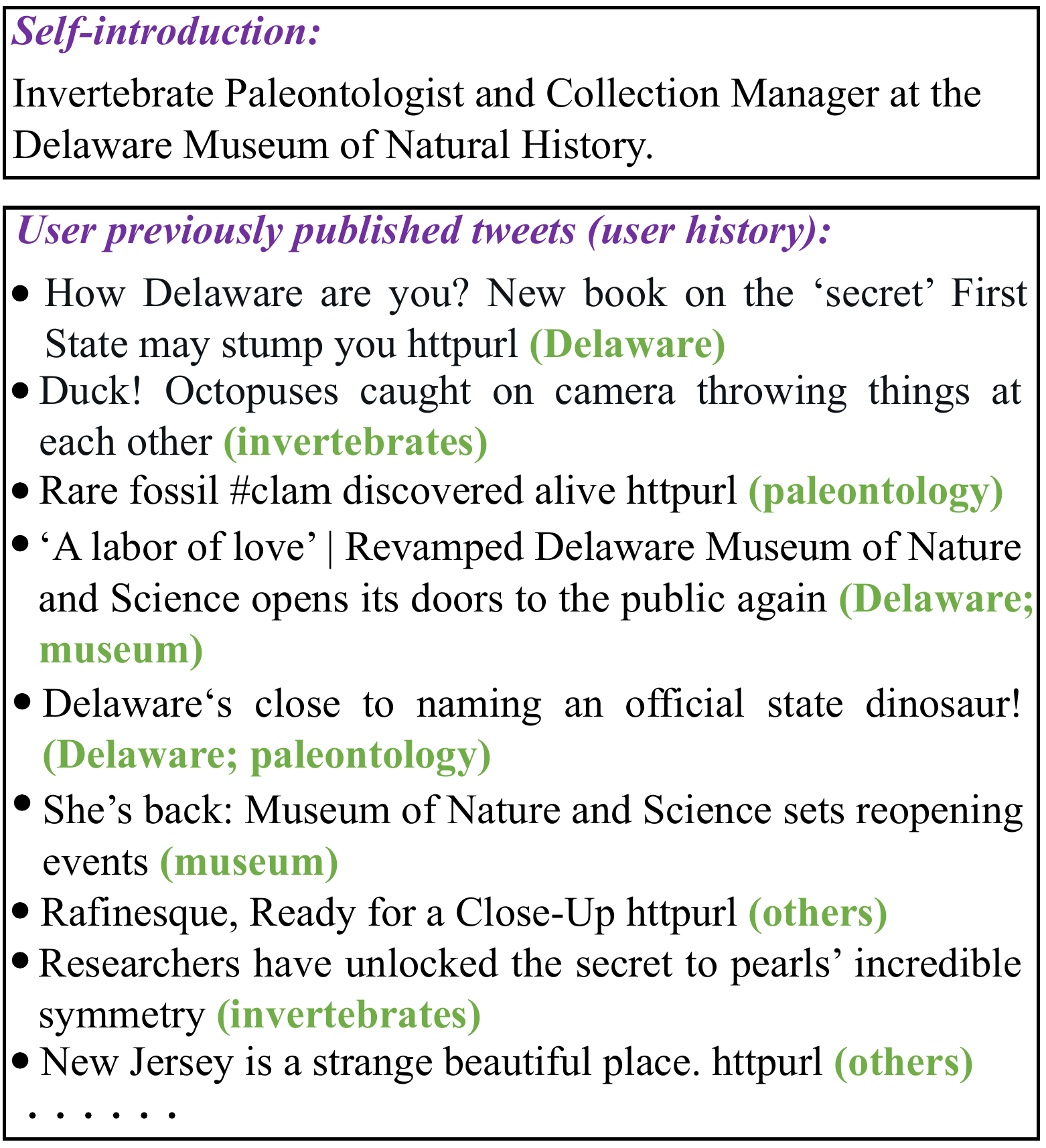}
\vspace{-2em}
\caption{Twitter user $U$ with a self-introduction on the top, followed by the previous tweets (user history).
$U$ exhibits a mixture of personal interests in
Delaware, invertebrates, paleontology, museum, and others.
}
 \vspace{-1.5em}
\label{fig:intro_case}
\end{figure}

%% file: Table/dataset_analysis.tex
\begin{table*}[t]
	 \centering
{\renewcommand{\arraystretch}{1.0}
\resizebox{1.9\columnwidth}{!}
{
	\begin{tabular}[b]{
	l crrrrr
	}
		\hline
		\multirow{2}*{\textbf{Datasets}}        & \multirow{2}*{\textbf{Data Source}} &  \multicolumn{3}{c}{\textbf{Source-Target Pair Number}} & \multicolumn{2}{c}{\textbf{Token Number}} \\
	\cmidrule(r){3-5} \cmidrule(r){6-7}
        & & Train & Valid & Test   &  Src. len. & Tgt. len.  \\
		\hline
        NYT \cite{NYT_dataset} & News & 44,382 &5,523 &6,495 &1183.2 &110.8 \\
        PubMed\cite{DBLP:conf/naacl/CohanDKBKCG18} & Scientific Paper & 83,233 & 4,946 & 5,025 & 444.0 & 209.5 \\
        Reddit \cite{DBLP:conf/naacl/KimKK19} & Social Media & 41,675 & 645 & 645 & 482.2 & 28.0 \\
        WikiHow \cite{DBLP:journals/corr/abs-1810-09305} & Knowledge Base & 168,126 & 6,000 & 6,000 & 580.8 & 62.6 \\
        Ours (\textit{users' self-introductions}) & Social Media & 140,956 & 17,619 & 17,624 & 1581.3 & 20.0 \\
		\hline	\end{tabular}}}
	\vspace{-0.5em}
	\caption{
	Statistical comparison of our social media self-introduction dataset with other popular summarization datasets. Src. means source (input), Tgt. refers to target (output), and len. stands for average length (word number). 
	}
	\vspace{-1em}
	\label{tab:dataset_analysis}
\end{table*}


%% file: Sections/Related_Work.tex
\section{Related Work\label{sec:related-work}}

Our work relates to user profiling (by task formulation) and topic modeling (by methodology). 

\paragraph{User Profiling.} 
This task aims to 
characterize user attributes to reflect a personal view.
Most previous work focuses on modeling a user's tweeting history \cite{DBLP:conf/acl/LiRH14} and social network interactions  \cite{DBLP:conf/naacl/QianSJGB19, DBLP:conf/kdd/WangFXL19, DBLP:conf/ijcai/ChenGRHXGYZ19, DBLP:conf/aaai/WangW0ZHF21, DBLP:conf/wsdm/WeiHXXZY22} to predict user attribute tags (e.g., ages and interests).
However, most existing work focuses on classifying user profiles into 
fragmented and limited tags.
Different from them, we study sentence-level self-introduction and explore how NLP handles such personalized generation, which initializes the potential to profile a user via self-introduction writing.

\paragraph{Topic Modeling.} Topic models are popular unsupervised learning methods to explore corpus-level word co-occurrence statistics and represent latent topics via clustering topic-related words. 
Recent work mostly adopts neural architectures based on Variational Autoencoder (VAE) \cite{DBLP:journals/corr/KingmaW13}, enabling easy joint work with other neural modules \cite{DBLP:conf/iclr/SrivastavaS17}. 

Latent topics have shown beneficial to many NLP writing applications, such as the language generation for dialogue summaries \cite{ DBLP:conf/aaai/0001ZZA21}, dialogue responses \cite{DBLP:conf/acl/ZhaoZE17, DBLP:conf/acl/EskenaziLZ18, DBLP:conf/acl/ChanLLZZSY21,DBLP:conf/emnlp/WangCL22}, poetries \cite{DBLP:conf/ijcai/ChenYSLYG19, DBLP:conf/aaai/YiLYLS20}, social media keyphrases \cite{DBLP:conf/acl/WangLCKLS19}, quotations \cite{DBLP:conf/emnlp/WangLZZW20}, and stories 
\cite{DBLP:conf/naacl/HuYLSX22}. 
Most existing methods focus on exploiting topics in decoding and injecting latent topic vectors (topic mixture) to assist generation.
%
In contrast to the above work's scenarios,
our application requires digesting much more lengthy and noisy inputs with scattered keypoints; thus, we leverage topics more finely and enable its joint guidance in encoding (by feeding in the topic mixture as topic prompts) and decoding (using topic words to control word-by-word generation).


Inspired by the success of pre-trained language models (PLMs), some efforts have been made to incorporate PLMs into VAE to conduct topic modeling \cite{DBLP:conf/emnlp/LiGLPLZG20,DBLP:conf/naacl/GuptaCS21, DBLP:conf/www/0001ZHZ022}. However, PLMs might be suboptimal in modeling user history (formed by numerous noisy tweets), because PLMs tend to be limited in encoding very long documents   \cite{cao-wang-2022-hibrids}. 
Here, we model latent topics by word statistics, allowing better potential to encode long input. 



%% file: Sections/dataset_collection.tex
\section{\label{sec:dataset}Twitter Self-Introduction Dataset}

To set up empirical studies for social media self-introduction, we build a large-scale Twitter dataset. 

\input{figure_sections/score_dist_sent_sum_dist.tex}
\paragraph{Data Collection.} 
Following \citet{DBLP:conf/emnlp/NguyenVN20}, we first downloaded the general Twitter streams from September 2018 to September 2019. 
Then,  we extracted the user ids therein  and removed the duplicated ones.
Next, we gathered users' tweeting history and self-introductions
via Twitter API\footnote{\url{https://developer.twitter.com}} and filtered out inactive users with less than  30 published tweets. 
For users with over 100 published tweets, only the latest 100 ones were kept. 
At last, we maintained the tweet text in English and removed  irrelevant fields, e.g., images and videos. 


\paragraph{Data Pre-processing.}
First, we removed non-English self-introductions and those too short ($<$7 tokens) or too long ($>$30 tokens).
Second, 
we employed SimCSE \cite{DBLP:conf/emnlp/GaoYC21} (an advanced model for semantic matching)
to measure the text similarity between a user's self-introduction and their tweeting history. 
Then, for training quality concern, we removed users with self-introductions that exhibit less than 0.4 similarity score \footnote{The details of the tweets with less than 0.4 similarity score are shown in Appendix \ref{append:data_filtering}.} on average to the top-30 tweets in history.\footnote{Users may have less than 30 tweets kept in history (e.g., some tweets without English text were excluded). For these users, we considered all their maintained tweets.}
Third, for the remaining 176,199 unique user samples, each corresponds to a pair of user history (source) and self-introduction (target). 
For model evaluation, we randomly split the user samples into training (80$\%$), validation (10$\%$), and test (10$\%$) sets.

\paragraph{Data Analysis.}
Encoder-decoder models are widely used in summarization tasks ($\S$\ref{sec:related-work}).
We then discuss the difference of our task through an empirical lens.
The statistics of our dataset and other popular summarization datasets are compared in Table \ref{tab:dataset_analysis}.
We observe that each of our data sample exhibits a longer source text and a shorter target text compared to other datasets.
It indicates the challenge of our self-introduction task, where directly using summarization models may be ineffective.


To further analyze the challenges, Figure \ref{fig:score_sent_num}(a) displays the distribution of SimCSE-measured source-target similarity 
(averaged over top-30 tweets in user history).
It implies that very few tweets are semantically similar to their authors' self-introductions, making it insufficient to simply ``copy'' from history tweets.
We then analyze and show the tweet number distribution in user history in Figure \ref{fig:score_sent_num}(b). 
It is noticed that 37\% users posted over 90 history tweets, scattering interest points in numerous tweets and hindering models in capturing the essential ones to write a self-introduction.

%% file: figure_sections/score_dist_sent_sum_dist.tex
\begin{figure}[t]
\centering
\includegraphics[width=1\columnwidth]{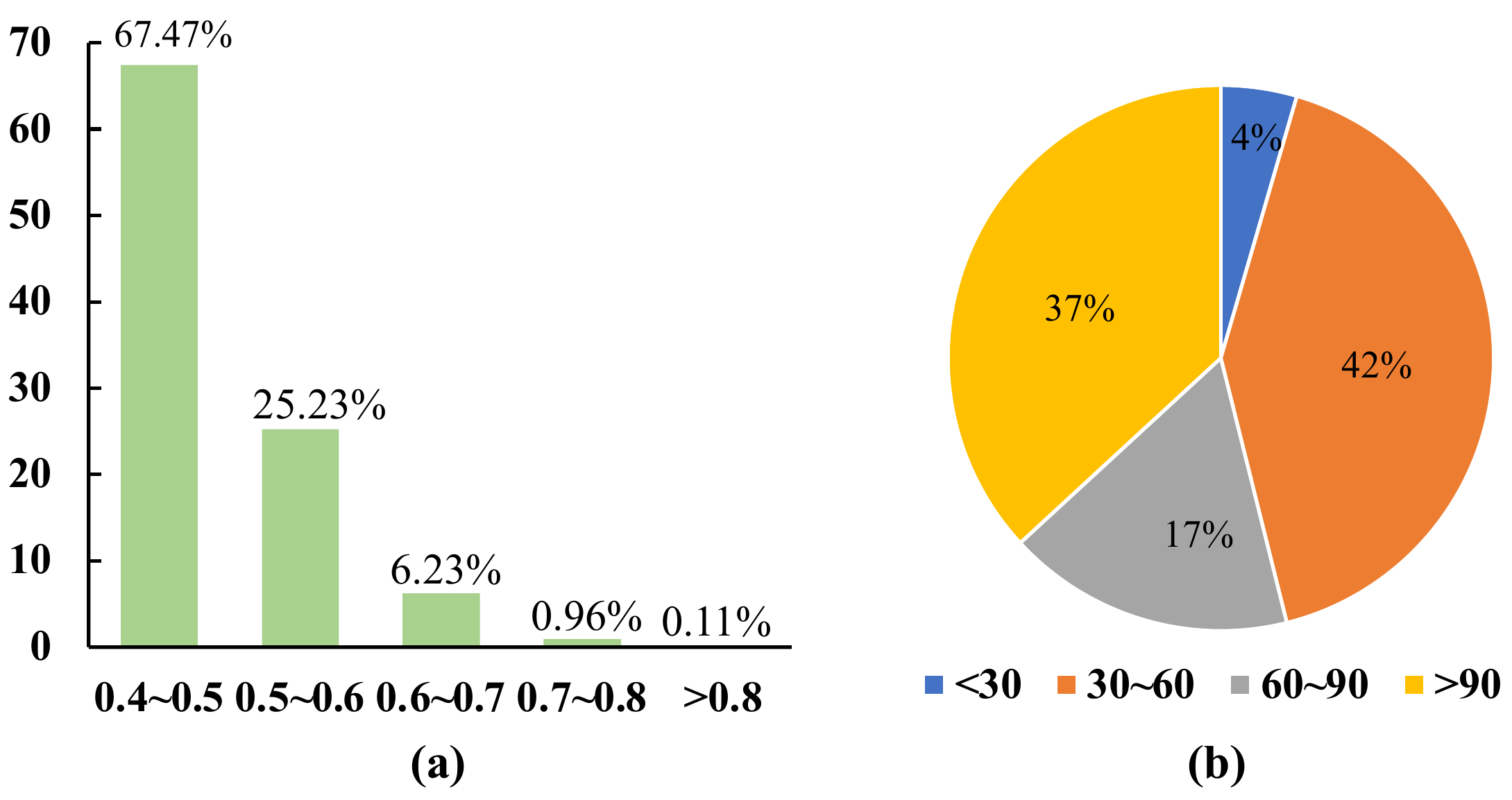}
\vspace{-2em}
\caption{\label{fig:score_sent_num}
Analysis of the distribution over (a)  average similarity of user history tweets (capped at top 30) to self-introduction and (b) tweet number in user history.
}
\vspace{-1em}
\end{figure}

%% file: Sections/method.tex
\section{\label{sec:model}Our UTGED Framework}
\input{figure_sections/model_architecture.tex}

Here we describe our UTGED (unified topic-guided encoder-decoder) framework. Its overview is in Figure \ref{fig:model}: latent topics guide the PLMs to encode user history and decode self-introductions. 

The data is formulated as source-target pairs $\{X^i,Y^i\}_{i=1}^{N}$, where $X^i=\{x^i_1,x^i_2,...,x^i_m\}$ indicates user history with $m$ tweets published by user $u^i$, $Y^i$ represents the user-written description, and $N$ is the number of pairs.
In our task, for user $u^i$, models are fed in their user history tweets $X^i$ and trained to generate their self-introduction $Y^i$. 

\subsection{Neural Topic Model}\label{ssec:model:NTM}
To explore users' interests hidden in their numerous and noisy tweets, we employ a neural topic model (NTM) \cite{DBLP:conf/iclr/SrivastavaS17} to learn latent topics (word clusters). NTM is based on VAE with an encoder and a decoder to reconstruct the input. 

For word statistic modeling, the history tweets in $X^i$ are first processed to a one-hot vector $X^i_{bow} \in \mathbb{R}^{V_{bow}}$ in Bag-of-words (BoW), where $V_{bow}$ indicates NTM's vocabulary size. 
Then, similar to VAE, NTM encoder transforms BoW vector $X^i_{bow}$ into a $K$-dimensional latent topic variable $z^i \in \mathbb{R}^K$.
Conditioned on $z^i$,
NTM decoder produces $\hat{X}^i_{bow}$
to reconstruct $X^i_{bow}$.
Here presents more details.

\paragraph{NTM Encoder.} 
Given the BoW vector $X^i_{bow}$, NTM encoder attempts to learn the mean $\mu^i$ and standard deviation $\sigma^i$ based on the assumption that words in $X^i$ exhibit a Gaussian prior distribution.
Its mean and standard deviation, $\mu^i$ and $\sigma^i$ will be encoded by the following formula and later be utilized to compute the 
latent topic vector $z^i$:
\vspace{-0.5em}
\begin{equation} \small
\mu^i = f_{\mu}(f_b(X^i_{bow})); \text{log} \sigma^i = f_{\sigma}(f_b(X^i_{bow}))
\vspace{-0.5em}
\end{equation}
where $f_{*}(\cdot)$ indicates a single layer perceptron performing the linear transformation of input vectors. 

\paragraph{NTM Decoder.} We then reconstruct the BoW in $X^i$ based on the NTM-encoded $\mu^{i}$ and $\sigma^{i}$. 
We hypothesize that a corpus may exist $K$ latent topics, each reflecting a certain user interest and represented by word distribution over the vocabulary $V_{bow}$.
Besides, user history $X^i$ is represented as a topic mixture $\theta^i$ to reflect $u^i$'s interest combination over $K$ topics. 
The procedure is as follows:

$\bullet$ Draw latent topic vector $z^i \sim \mathcal{N}(\mu^i, \sigma^i)$

$\bullet$ Topic mixture $\theta^i=\text{softmax}(f_{\theta}(z^i))$

$\bullet$ For each word $w \in X^i$: \\
$\indent$$\indent$$\indent$$\indent$Draw $w \sim \text{softmax}(f_{\phi}(\theta^i))$ \\
where $f_{\theta}$ and $f_{\phi}$ are a single layer perceptron.
The weight matrix of $f_{\phi}$ indicates topic-word distributions ($\phi_1$,$\phi_2$,...,$\phi_K$). 

The learned latent topics for $X^i$ will later
guide the BART-based self-introduction generation (to be discussed in $\S$\ref{ssec:model:topic_guided_model}).
The topic mixture $\theta^i$ will be injected into the BART encoder for capturing salient interests and 
the top-$l$ words $A^i=\{a^i_1, a^i_2,...,a^i_l\}$ with highest topic-word probability in $\phi_c$ ($c$ indexes the major topic suggested by $\theta^i$)
will go for controlling the writing process of the BART decoder. 

\subsection{Topic-Guided Generation Model}\label{ssec:model:topic_guided_model}
We have discussed how to model a user $u^i$'s latent interests with NTM and the learned latent topics ($\theta^i$ and topic words $A^i$) will then guide a BART-based encoder-decoder model to generate $u^i$'s self-introduction, $Y^i$. 
In the following, 
we first present how we select tweets (for fitting overly long user history into a transformer encoder), followed by our topic-guided design for encoding and decoding.

\paragraph{Tweet Selection.}
Recall from $\S$\ref{sec:dataset} that user history tends to be very long (Table \ref{tab:dataset_analysis} shows it has 1581.3 tokens on average). However, BART encoder limits its input length. 
To fit in the input, we go through the following steps to shortlist representative tweets from a user $u^i$'s lengthy tweeting history, $X^i$.

First, we measure how well a tweet $x_u^i$ can represent $X^i$ via averaging its  similarity to all others:

\vspace{-0.5em}
\begin{equation}\small\label{eq:sentence_similarity}
s^i_u=\frac{1}{|m|}\sum_{x^i_v \in X^i } Sim\left ( x^i_u,x^i_v \right )
\vspace{-0.5em}
\end{equation}
where $ Sim(x^i_u,x^i_v)$ represents the $x^i_u - x^i_v$ SimCSE-measured cosine similarity. 
Then, we maintain a shortlist $R_i$ to hold $X^i$'s representative tweets, which is empty at the beginning and iteratively added with $x^i_h$ obtaining the highest similarity score (Eq. \ref{eq:sentence_similarity}). To mitigate redundancy in $R^i$, once $x^i_h$ is put in $R_i$, it is removed from $X^i$, and so are other tweets in $X^i$ whose cosine similarity to $x^i_h$  is over a threshold $\lambda$ (i.e., 0.8). 
For easy reading, we summarize the above steps in 
Algorithm  \ref{alg:sent_filter}.

After that, we further rank the shortlisted tweets in $R^i$ based on their overall similarity in $X^i$ (Eq. \ref{eq:sentence_similarity}).
The top ones are maintained and concatenated chronologically to form a word sequence $R^i=\{w^i_{1},w^i_{2},...,w^i_{M}\}$ ($M$ denotes the word number).

\paragraph{Topic Prompt Enhanced Encoder (TPEE).} 
We then discuss how we encode $R^i$ (selected user history tweets) in guidance of the topic mixture $\theta^i$ (featuring latent user interests).
The encoding adopts the BART encoder and is trained with $\theta^i$-based prompt fine-tuning (thereby named as TPEE, short for topic prompt enhanced encoder).

We first obtain the topic prompt as follows:
\vspace{-0.5em}
\begin{equation} \small\label{eq:prompt}
B^i = \text{MLP} (\theta^i)
\vspace{-0.5em}
\end{equation}
where $\text{MLP}$ is a feedforward neural network. 
Following \citet{DBLP:conf/acl/LiL20}, $B^i \in \mathbb{R}^{d \times L}$ is split into $L$ vectors $[b^i_1,b^i_2,...,b^i_L]$. $L$ indicates the topic prompt length and each vector $b^i_j \in \mathbb{R}^d$.

To inject the guidance from topic prompts $\{b^i_1,b^i_2,...,b^i_{L}\}$ (carrying latent topic features), we put them side by side with the embeddings of words $\{w^i_{1},w^i_{2},...,w^i_{M}\}$ (reflecting word semantics of $R^i$). 
Then, a BART encoder $\mathcal{E}$ represents 
user $u^i$'s salient interests  $H^i_E$ in its last layer: 
\vspace{-0.5em}
\begin{equation} \small\label{eq:encoder}
H^i_E = \mathcal{E}(\left [b^i_1;b^i_2;...;b^i_{L}; e^i_1;e^i_2;...;e^i_M\right ])
\vspace{-0.5em}
\end{equation}
where $e^i_j \in \mathbb{R}^d$ is the BART-encoded word embedding of
$w^i_j$ and 
$\left[; \right]$ is the concatenation operation.

\paragraph{Topic Words Enhanced Decoder (TWED).}
Recall in $\S$\ref{ssec:model:NTM}, NTM generates $l$ topic words ($A^i$) to depict a user $u^i$'s major latent interests.
To further reflect such interests in the produced self-introduction, we employ $A^i$ to control a BART decoder $\mathcal{D}$ in its word-by-word generation process through the topic control module.

For easy understanding, we first describe how the original BART decode.
At the $t$-th step, the decoder $\mathcal{D}$ is fed in its previous  hidden states $H^i_{D,t}$, the BART encoder's hidden states $H^i_E$ (Eq. \ref{eq:encoder}), and latest generated word $Y^i_t$, resulting in hidden step $o^i_{t+1}$.
Based on that, the next word is generated following the token distribution $p_{t+1}^i$. The concrete workflow is shown in the formula as follows:
\vspace{-0.5em}
\begin{equation} \small \label{eq:token_predict}
p^i_{t+1}=\text{softmax}(W_{e} o^i_{t+1})
\vspace{-0.5em}
\end{equation}
\vspace{-1em}
\begin{equation} \small \label{eq:decoder}
o^i_{t+1}, H^i_{D,t+1} = \mathcal{D}(H^i_E,H^i_{D,t}, Y^i_t)
\vspace{-0.5em}
\end{equation}
where 
$H^i_{D,t+1}$ stores all the previous decoder hidden states till step $t+1$,
$W_{e}$ is learnable and to map the latent logit vector $o_{t+1}^i$ to the target vocabulary. 

Then, we engage topic words $A^i$ to control the above procedure by the topic control module.
Inspired by BoW attribute model \cite{DBLP:conf/iclr/DathathriMLHFMY20},
we calculate the following log-likelihood loss to weigh the word generation probability $p_{t+1}^i$ over each topic word $a_j^i\in A^i$:
\vspace{-0.5em}
\begin{equation} \small \label{eq:atribute_loss}
\log{p(A^i|Y^i_{t+1})} = \log{(\sum_{j}p^i_{t+1}[a^i_j])}
\vspace{-0.5em}
\end{equation}

The gradient from $\log{p(A^i|Y^i_{t+1})}$ is further involved in updating all decoder layers ($H^i_{D,t}$) in $\mathcal{D}$:
\vspace{-0.5em}
 \begin{equation}\small \label{eq:updated_result}
 \widetilde{H}^i_{D,t} = \Delta H^i_{D,t}+H^i_{D,t}
 \end{equation}
\begin{equation}\small  \label{eq:updated_state}
    \Delta H^i_{D,t} \leftarrow \Delta H^{i}_{D,t}+\alpha\frac{\nabla_{\Delta H^{i}_{D,t}}\log{p(A^i|H^i_{D,t}+\Delta H^i_{D,t})}}{\lVert\nabla_{\Delta H^i_{D,t}}\log{p(A^i|H^i_{D,t}+\Delta H^i_{D,t})}\rVert ^{\gamma}}
\end{equation}
where $\widetilde{H}^i_{D,t}$ indicates the updated (topic-controlled) decoder's states, $\Delta H^i_{D,t}$ means the gradient update to $H^i_{D,t}$, $\alpha$ is the step size, and $\gamma$ is the normalization value. 
Furthermore, we adopt the same topic-controlling strategy to update the encoder's final layer states $H^i_E$ and derive the updated states $\widetilde{H}^i_E$ based on Eq. \ref{eq:updated_result} and \ref{eq:updated_state}. With Eq. \ref{eq:token_predict} and \ref{eq:decoder}, we can accordingly obtain the final token distribution $\widetilde p^i_{t+1}$ based on the topic-controlled encoder and decoder states $\widetilde{H}^i_E$, $\widetilde{H}^i_{D,t}$, and previous predicted word $Y^i_t$.

\subsection{Joint Training in a Unified Framework}\label{ssec:model:objectives}
To couple the effects of NTM (described in $\S$\ref{ssec:model:NTM}) and topic-guided encoder-decoder module for self-introduction generation  (henceforth SIG discussed in $\S$\ref{ssec:model:topic_guided_model}), we explore the two modules in a unified framework and jointly train them for better collaborations.
The loss function of the unified framework is hence a weighted sum of NTM and SIG:

\vspace{-0.5em}
\begin{equation}\small \label{eq:loss_function}
 \mathcal{L}=\alpha \mathcal{L}_{NTM}+(1-\alpha) \mathcal{L}_{SIG}
\end{equation}
where $\mathcal{L}_{NTM}$ and $\mathcal{L}_{SIG}$  are the loss functions of NTM and SIG. $\alpha$ is the hyper-parameter trading off their effects and is set to 0.01 in our experiments.

For NTM, the learning objective is computed as:
\vspace{-0.5em}
\begin{equation}\small \label{eq:vae_loss}
 \mathcal{L}_{NTM}=D_{KL}({\delta}(z)|| {\rho} (z|X))-\mathbb{E}_{{\rho} (z|X)} [{\delta}(X|z)]
\end{equation}
where $D_{KL}(\cdot)$ indicates the Kullback-Leibler divergence loss and $\mathbb{E}[\cdot]$ is reconstruction loss.\footnote{We refer readers to more details of NTM in \citet{DBLP:conf/iclr/SrivastavaS17}, which are beyond the scope of this paper.}
For the SIG, it is trained with the cross-entropy loss:
\vspace{-0.5em}
\begin{equation}\small
 \mathcal{L}_{SIG}=-\sum_{i}\sum_{t} \log{p^{i}_{t}}
 \vspace{-0.5em}
\end{equation}

In practice, we first train the unified framework with Eq.\ref{eq:loss_function} and exclude $A^i$ (topic words output of NTM).
Then, during inference, we fix UTGED, employ $A^i$ to control the decoding process and generate the final self-introduction with Eq.\ref{eq:atribute_loss}\textasciitilde Eq.\ref{eq:updated_state}.

%% file: figure_sections/model_architecture.tex
\begin{figure}[t]
\centering
\includegraphics[width=1\columnwidth]{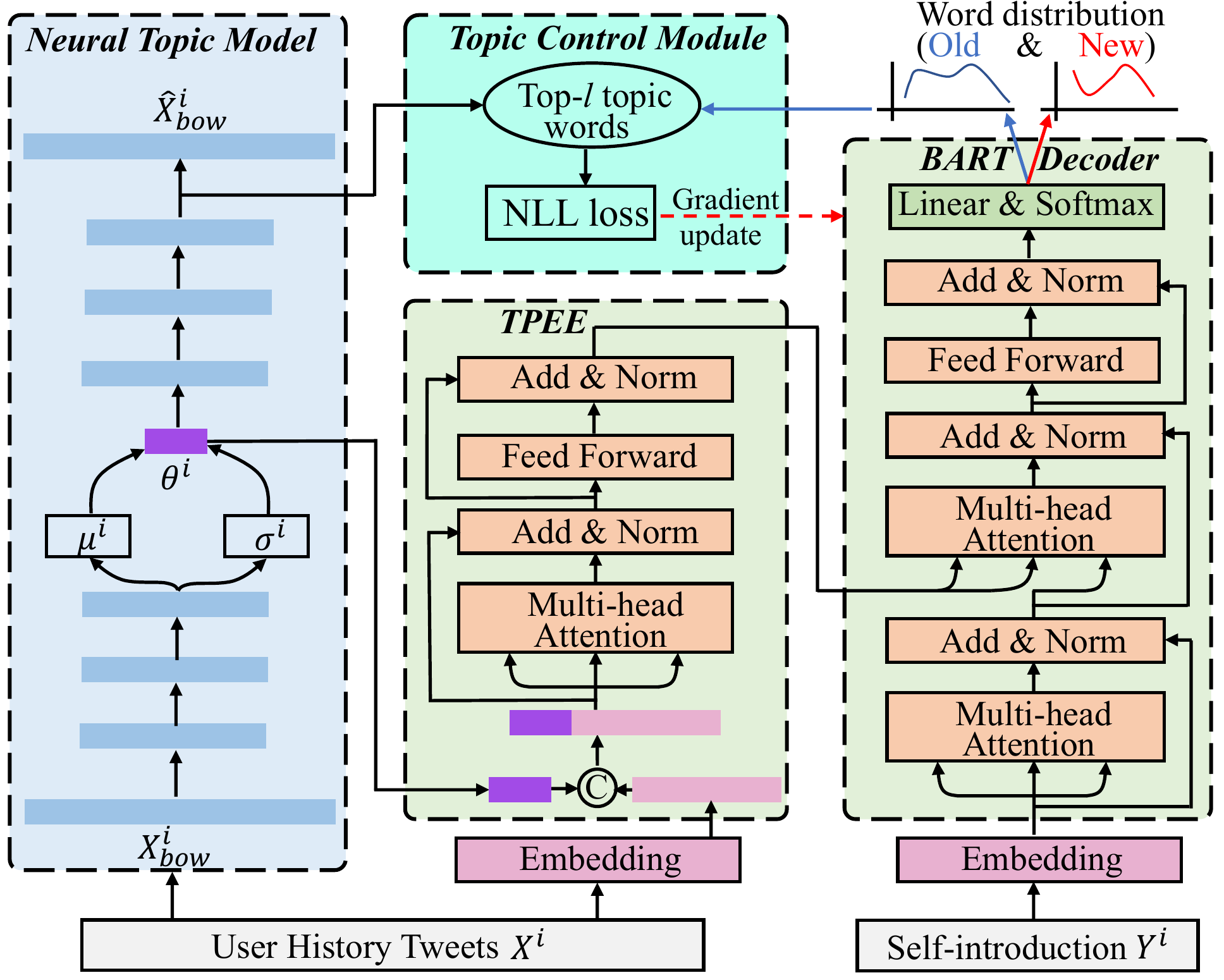}
\vspace{-1.5em}
\caption{\label{fig:model} The overview of our UTGED (Unified Topic-Guided Encoder-Decoder) framework, 
The left module shows a neural topic model (NTM) representing user interests with latent topics. The topic mixtures help the encoder explore user history (middle) and topic words guide the  decoder in self-introduction generation (right). 
}
\vspace{-1em}
\end{figure}

%% file: Sections/Experimental_Setup.tex
\section{Experiments and Discussions}

\subsection{Experimental Setup}

\paragraph{Model Settings.} 
We implemented NTM ($\S$\ref{ssec:model:NTM}) based on \cite{DBLP:conf/iclr/SrivastavaS17} and set its topic number $K$ to 100. 
Its BoW vocabulary size $V_{bow}$ is set to 10K and hidden size to 200. 
The input of NTM is the BoW of original user history $X^i$ while the input of SIG is capped at 1,024 tokens based on the shortlisted tweets in $R^i$ ($\S$\ref{ssec:model:topic_guided_model}).\footnote{We also test NTM with BoW on $\{R^i\}_{i=1}^{N}$ and observe slightly worse results. It is possibly because NTM is based on word statistics and would barely be affected by lengthy input.} 

The SIG model is based on the BART and built on 6 encoding layers and 6 decoding layers.
We adopted AdamW and SGD to optimize the SIG and NTM, respectively. 
The learning rate is set to $5\times 10^{-5}$ for SIG and $1\times 10^{-4}$ for NTM. The topic prompt length $L$ is set to 7.
To warm up joint training (Eq.\ref{eq:loss_function}), we pre-train NTM with Eq.\ref{eq:vae_loss} for 100 epochs. During joint training, batch size is set to 8 and the maximum epoch to 5.
In topic-controlled decoding, $\alpha$ is set to 0.25 and $\gamma$ to 1.5 (Eq.\ref{eq:updated_state}). Topic word number $l$ is set to 30.
Models are trained on a 24GB NVIDIA RTX3090 GPU. 

\paragraph{Evaluation Metrics.}
For automatic evaluation, we adopt 
ROUGE-1 (R-1), ROUGE-2 (R-2), and ROUGE-L (R-L), which are popular metrics in language generation based on output-reference word overlap and originally for summarization tasks \cite{lin-2004-rouge}.
We also conduct a human evaluation on a 5 point Likert scale and over three criteria: \textit{fluency} of
the generated language,  \textit{consistency} of a self-introduction to the user's history, and  \textit{informativeness} of it to reflect essential user interests. 




\paragraph{Baselines and Comparisons.} 
We adopt extractive and abstractive summarization models in comparison.
The former extracts user history tweets as the self-introduction by ranking them with: (1)
\textbf{BERTExt} \cite{DBLP:conf/emnlp/LiuL19} (based on BERT \cite{DBLP:conf/naacl/DevlinCLT19})
(2) \textbf{TextRank} \cite{DBLP:conf/emnlp/MihalceaT04} (unsupervised graph ranking based on similarity)
(3) \textbf{Consen} ( unsupervised ranking with the averaged similarity to others (Eq.\ref{eq:sentence_similarity})). 

For abstractive competitors, models all follow the encoder-decoder paradigm.
We employ \textbf{T5} \cite{DBLP:journals/jmlr/RaffelSRLNMZLL20}, \textbf{BART} \cite{DBLP:conf/acl/LewisLGGMLSZ20}, and \textbf{PEGASUSU-X} \cite{DBLP:journals/corr/abs-2208-04347}, all based on PLMs and are state-of-the-art abstractive summarizers. 
We also compare to \textbf{GSum} \cite{DBLP:conf/naacl/DouLHJN21},
which employs highlighted sentences, keywords, relations, and retrieved summaries.

In addition, we examine the upper-bound tweet selection (shortlist given reference self-introduction).
Here SimCSE first measures the similarity between the reference and each tweet in user history $X^i$. 
$\textbf{Oracle}_{\textbf{E}}$ then extracts the tweet with the highest similarity score.
For $\textbf{Oracle}_{\textbf{A}}$, we rank tweets based on the similarity score and the top ones are fed into BART for a generation. 
Furthermore, to explore the potential of our topic-guided design over $\textbf{Oracle}_{\textbf{A}}$ model, 
we feed $\textbf{Oracle}_{\textbf{A}}$'s input to our UTGED and name it $\textbf{Oracle}_{\textbf{A}}\textbf{+Topic}$.

%% file: Sections/Experimental_Results.tex
\input{Table/main_results.tex}
\input{Table/ablation_results.tex}





\subsection{Main Comparison Results} \label{ssec:exp:main}
Table \ref{tab:main_results} shows the main comparison results.
We first observe the inferior results from all extractive models, including
$\text{Oracle}_{E}$.
It is because of the non-trivial content gap between users' history tweets and their self-introductions (also indicated in Figure \ref{fig:score_sent_num}). 
Directly extracting tweets from user history is thus infeasible to depict self-introduction, presenting the need to involve language generation. 
For this reason, abstractive methods exhibit much better performance than extractive baselines.



Among comparisons in abstractive models, UTGED yields the best ROUGE scores and significantly outperforms the previous state-of-the-art summarization models.
It shows the effectiveness in engaging guidance of latent topics from lengthy and noisy user history, which may usefully signal the salient interests for writing a self-introduction.


In addition, by comparing $\text{Oracle}_{\text{A}}$  results and model results, we observe a large margin in between. 
It suggests the challenge and importance of tweet selection for user history encoding, providing insight into future related work.
Moreover, interestingly, $\text{Oracle}_{\text{A}}\text{+Topic}$ further outperforms $\text{Oracle}_{\text{A}}$, implying topic-guided design would likewise benefit the upper-bound tweet selection scenarios.


\paragraph{Ablation Study.}
Here we probe into how UTGED's different modules work  and show the ablation study results in
Table \ref{tab:ablation_results}. 
All modules (tweet selection (S), TPEE (E), and TWED (D)) contrite positively because they are all designed to guide models in focusing on essential content reflecting user interests against lengthy input.
TPEE may show larger individual gain than the other two, possibly because the topic mixtures directly reflect  user interests and are easier for the model to leverage.

\input{Table/human_evaluation.tex}

\input{figure_sections/quantitative_analysis_parameter.tex}
\input{figure_sections/sent_num.tex}

\paragraph{Human Evaluation.}
To further test how useful our output is to human readers,  
we randomly select 100 samples from test set and train 3 in-house annotators from NLP background 
to rate the generated self-introductions. 
As shown in Table \ref{tab:human_eval}, UTGED is superior in informativeness and consistency. It implies latent topics can usefully help capture salient interests from lengthy and noisy user history.
However, its fluency is lower than that of BART, indicating that topic words slightly perturb the pre-trained decoder
\cite{DBLP:conf/iclr/DathathriMLHFMY20}.


\subsection{Quantitative Analysis}\label{ssec:exp:quantitative}
To better study UTGED, we then quantify the topic number, prompt length, and input tweet number to examine how they affect performance.
Here only R-L is shown for better display, and similar trends were observed from R-1 and R-2.
For the full results, we refer readers to Appendix \ref{append:full_results}.

\paragraph{Varying Topic Number.}
The first parameter analysis concerns 
the topic number $K$ (NTM's hyper-parameter).
As shown in Figure \ref{fig:all_parameter}(a), the score first increases then decreases with larger $K$ and peaks the results at $K=100$.
We also observe $K=200$ results in much worse performance than other $K$s, probably because modeling too fine-grained topics is likely to overfit NTM in user interest modeling, further hindering self-introduction generation. 

\paragraph{Varying Prompt length.} 
Likewise, we analyze the effects of prompt length $L$ in Figure \ref{fig:all_parameter}(b).
The best score is observed given $L$=7, much better than very short or very long prompt length.
Longer prompts may allow stronger hints from NTM, helpful to some extent; however, if the hint becomes too strong (given too-long prompt), topic features may overwhelm the encoder in learning specific features for self-introduction writing.  


\paragraph{Users w/ Varying Tweet Number.}
Recall in Figure \ref{fig:score_sent_num}(b), users largely vary tweet number in history (attributed to different active degrees).
We then examine how models work given varying tweet numbers in history. BART+S and UTGED are tested, both with tweet selection (S) to allow very long input and 
Figure \ref{fig:sent_num} shows the results. Both models exhibit growing trends for more active users, benefiting from richer content in their history to infer self-introduction.
Comparing the two models, UTGED performs consistently better, showing the gain from NTM's is robust over varying users. 


\subsection{Qualitative Analysis}\label{ssec:exp:qualitative}
\paragraph{Case Study.}
Figure \ref{fig:case_study} shows a user sample interested in ``teaching'' and ``reading''.
It can be indicated by topic words like ``student'', ``book'', and ``school'' produced by NTM.
From BART's output, we find its errors in ``seesaw specialist'' further mislead the model in writing more irrelevant content (e.g., ``google certified educator'' and ``google trainer''). 
It may be caused by the common exposure bias problem in language generation \cite{DBLP:journals/corr/RanzatoCAZ15, DBLP:conf/acl/ZhangFMYL19}.
On the contrary, UTGED's output is on-topic till the end, showing topic guidance may mitigate off-topic writing. \footnote{More topic word cases could be found in Appendix \ref{append:topic2words} and longer source tweets are shown in Appendix \ref{append:longer_case}.}


\paragraph{Error Analysis.}
In the main comparison (Table \ref{tab:main_results}), UTGED performs the best in yet also has a non-trivial gap to Orcation$_A$.
Here we probe its limitations and discuss the two major error types in Figure \ref{fig:error}. 
First, the output may contain grammatical mistakes, e.g., ``deals, deals", limited by BART'S decoder capability  and topic words' effects.
It calls for involving grammar checking in decoding. 
The second error type is propagated from wrong latent topics. 
As shown in the error case (second row), the user is a provider of ``pet''-style clothes, whereas NTM may cluster it with other ``pet lover''-users and further mislead the writing process. 
Future work may explore a better topic modeling method to mitigate the effects of mistakenly clustering.

\input{figure_sections/case_study.tex}
\input{figure_sections/error_analysis.tex}

%% file: Table/main_results.tex
\begin{table}[t]
	 \centering
{\renewcommand{\arraystretch}{0.9}
\resizebox{0.9\columnwidth}{!}
{
	\begin{tabular}[b]{l c c c}
		\toprule
		\textbf{Method}        & \textbf{R-1} & \textbf{R-2} &\textbf{R-L}  \\
		\midrule
		\textbf{\textit{Extractive}} \\
		\text{BERTExt} 
		& 11.67   & 1.92  & 10.04  \\
		\text{TextRank}    
		& 13.60   & 2.93  & 11.66  \\
		\text{Consen}    
		& 14.86   & 2.90  & 12.89   \\
            
		\hline
		\textbf{\textit{Abstractive}} \\
	    \text{T5} & 23.93  & 7.31   & 20.93  \\
	    \text{PEGASUS-X} & 24.10  & 7.44  & 21.07   \\
	    \text{GSum} & 22.19  & 5.99   & 19.27   \\
     	\text{BART} & 23.92  & 7.46   & 20.91   \\
      \text{UTGED (Ours)} & \textbf{24.99*}  & \textbf{8.05*}   & \textbf{21.84*}   \\
      \hline
      $\text{Oracle}_{E}$
            & 20.89   & 5.94  & 18.04   \\
            $\text{Oracle}_{A}$
            & 28.97  & 10.23   & 25.29   \\
            $\text{Oracle}_{A}$+Topic
            & 29.36  & 10.39   & 25.62   \\
		\bottomrule	\end{tabular}}}
	\vspace{-0.5em}
	\caption{
	Main comparison results. UTGED achieves the best results (highlighted) and the performance gain is significant to all comparison models (indicated by * and measured by paired t-test with p-value$<$0.05).
	}
	\label{tab:main_results}
\end{table}

%% file: Table/ablation_results.tex
\begin{table}[t]
	 \centering
{\renewcommand{\arraystretch}{0.9}
\resizebox{1.0\columnwidth}{!}
{
	\begin{tabular}[b]{l c c c}
		\toprule
		\textbf{Method}        & \textbf{R-1} & \textbf{R-2} &\textbf{R-L}  \\
		\midrule
		
     	\text{BART} & 23.92  & 7.46   & 20.91   \\
      \textsc{BART+S} & 24.26  & 7.68   & 21.17   \\
            \textsc{BART+S+E} & 24.78  & 7.95   & 21.65   \\
            \textsc{BART+S+E+D (UTGED)} & 24.99  & 8.05   & 21.84   \\

		\bottomrule	\end{tabular}}}
	\vspace{-2em}
	\caption{\label{tab:ablation_results}
	Ablation study results. S: tweet selection (to shortlist tweets from user history); E: w/ TPEE (topic-guided encoder); D: w/TWED (topic-guided decoder). 
	}
	\vspace{-1em}
\end{table}

%% file: Table/human_evaluation.tex
\begin{table}[t]
	 \centering
{\renewcommand{\arraystretch}{1.0}
\resizebox{1.0\columnwidth}{!}
{
	\begin{tabular}[b]{l c c c}
		\toprule
		\textbf{Method}        & \textbf{Fluency} & \textbf{Informativeness} &\textbf{Consistency}  \\
		\midrule
		
     	\text{GSum} & 3.43    & 2.65   & 2.28   \\
            \text{BART} & 3.85  & 3.21   &  2.89  \\
            \text{UTGED} & 3.66 & 3.68  &  3.27 \\

		\bottomrule	\end{tabular}}}
	\vspace{-2em}
	\caption{\label{tab:human_eval}
	Human evaluation results. Cohen’s Kappa for all annotator pairs is
0.63 on average
(good agreement).
	}
	\vspace{-0.5em}
	
\end{table}

%% file: figure_sections/quantitative_analysis_parameter.tex
\begin{figure}[t]
\centering
\includegraphics[width=1\columnwidth]{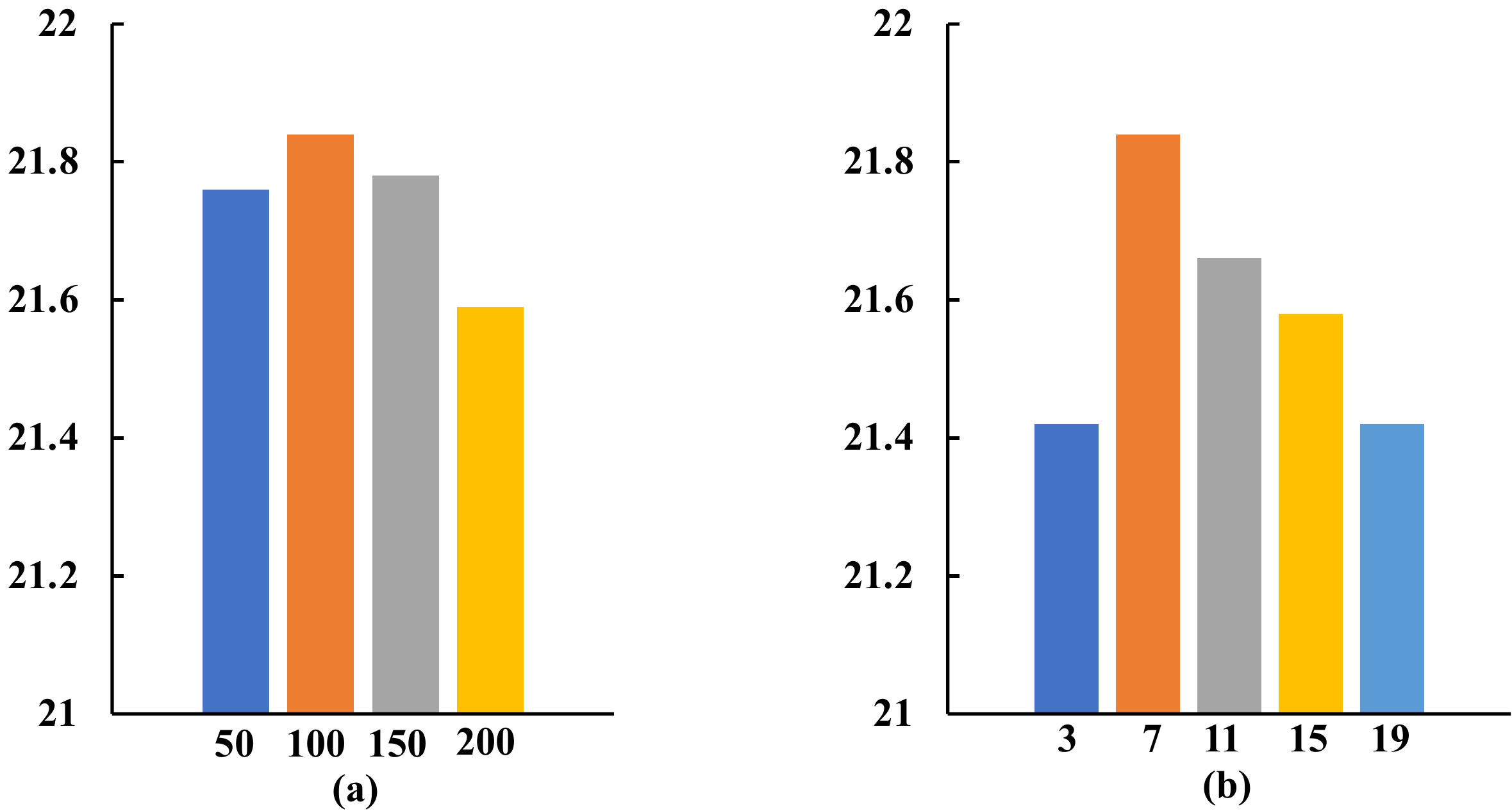}
\vspace{-2em}
\caption{\label{fig:all_parameter} Parameter analysis results. The X-axis shows topic number (a) and prompt length (b); Y-axix shows the R-L score measured on our UTGED's output.
}
\vspace{-1em}
\label{fig:all_parameter}
\end{figure}

%% file: figure_sections/sent_num.tex
\begin{figure}[t]
\centering
\includegraphics[width=1\columnwidth]{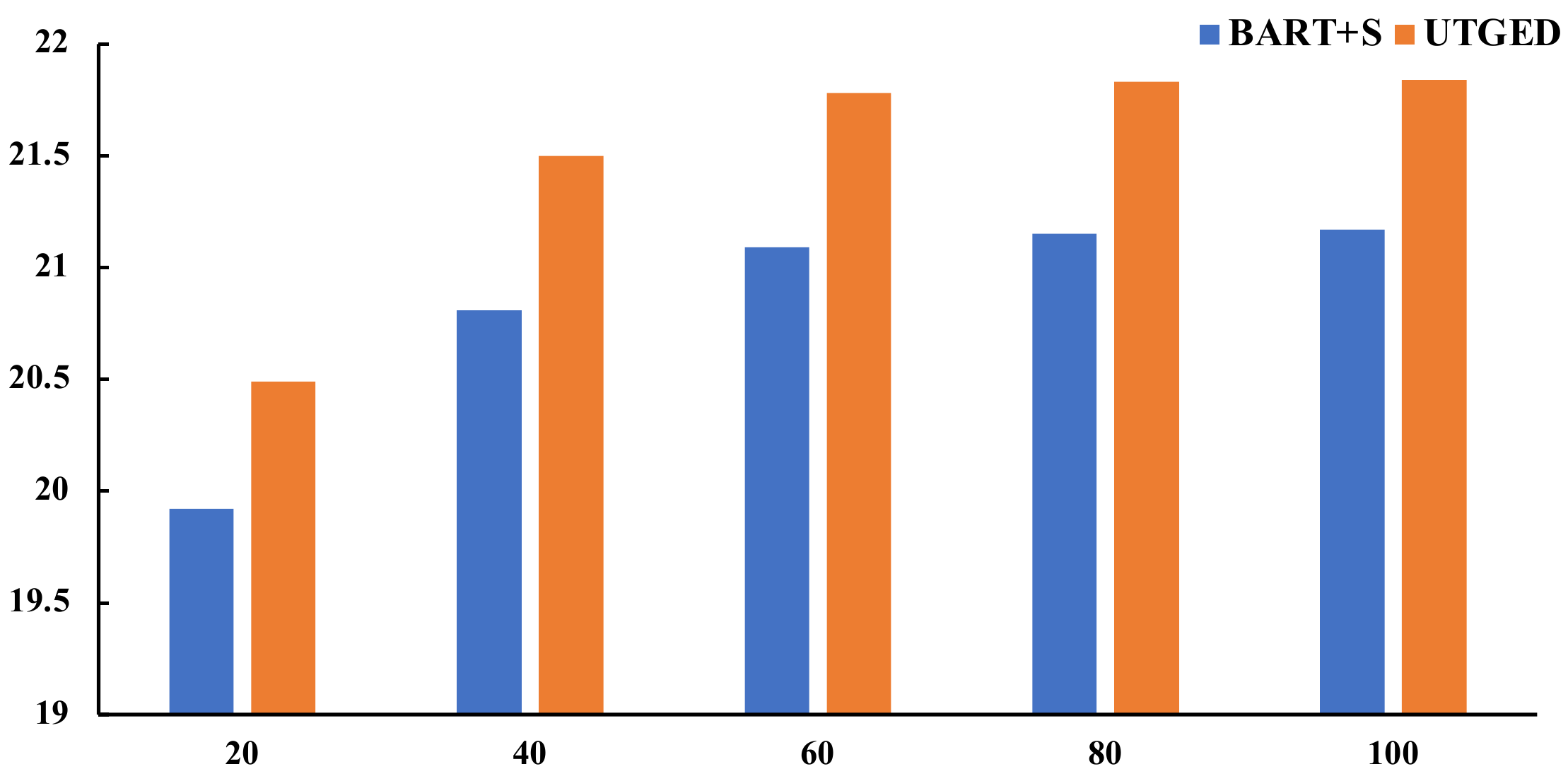}
\vspace{-2em}
\caption{\label{fig:sent_num} The comparisons between BART+S and UTGED while varing sentence number. X-axis: the values of sentence number; Y-axis: the R-L score.
}
\vspace{-1em}
\label{fig:sent_num}
\end{figure}

%% file: figure_sections/case_study.tex
\begin{figure}[t]
\centering
\includegraphics[width=1\columnwidth]{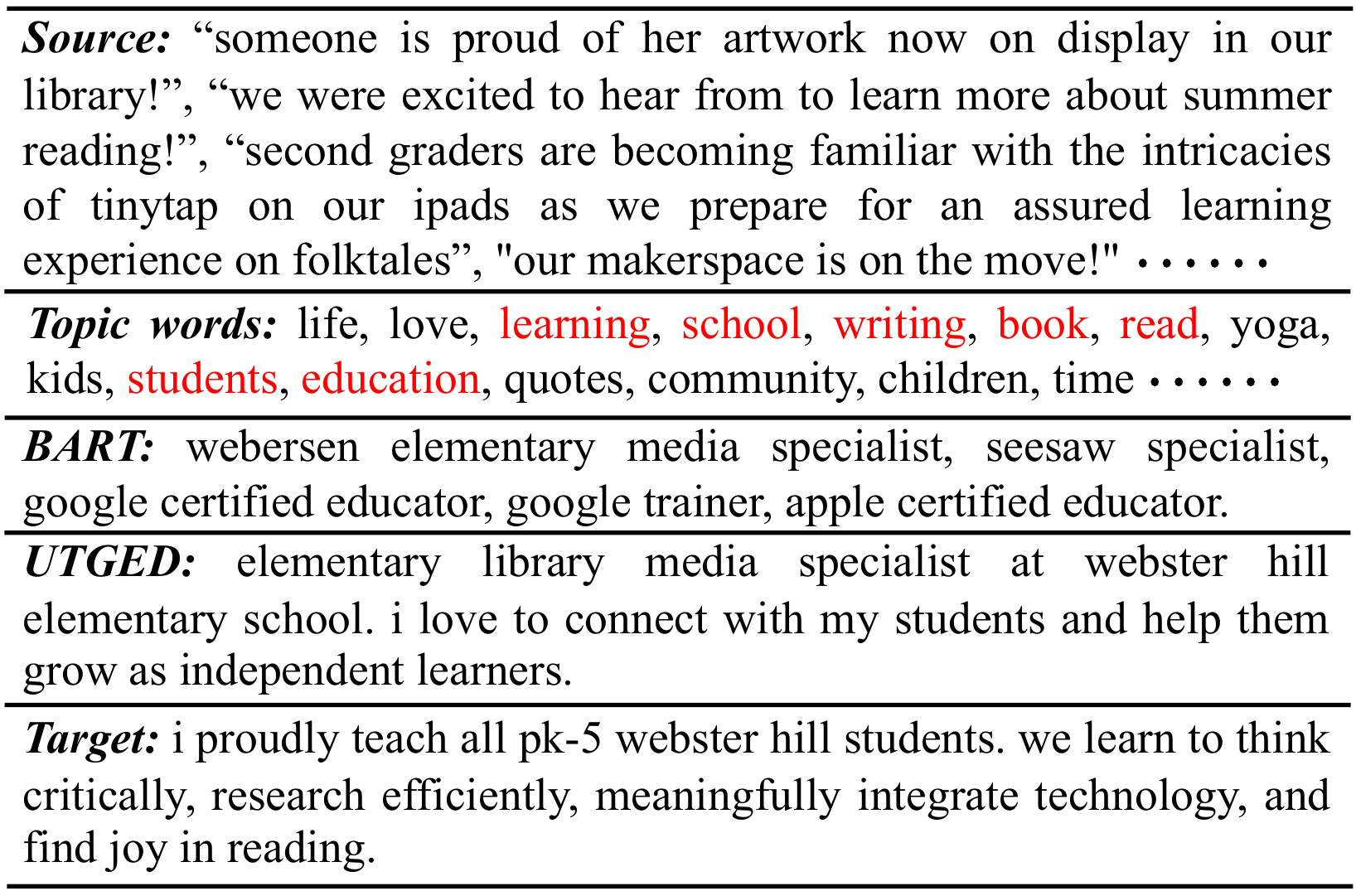}
\vspace{-1.75em}
\caption{
A Twitter user sample and the related results. 
From top to down shows user history (source $T^i$), major topic words ($A^i$), BART output, UTGED output, and reference self-introduction (target $Y^i$).
We inspect the topic words helpful for our task and color them in 
\textcolor{red}{red}. 
}
\label{fig:case_study}
\end{figure}

%% file: figure_sections/error_analysis.tex
\begin{figure}[t]
\centering
\includegraphics[width=1.0\columnwidth]{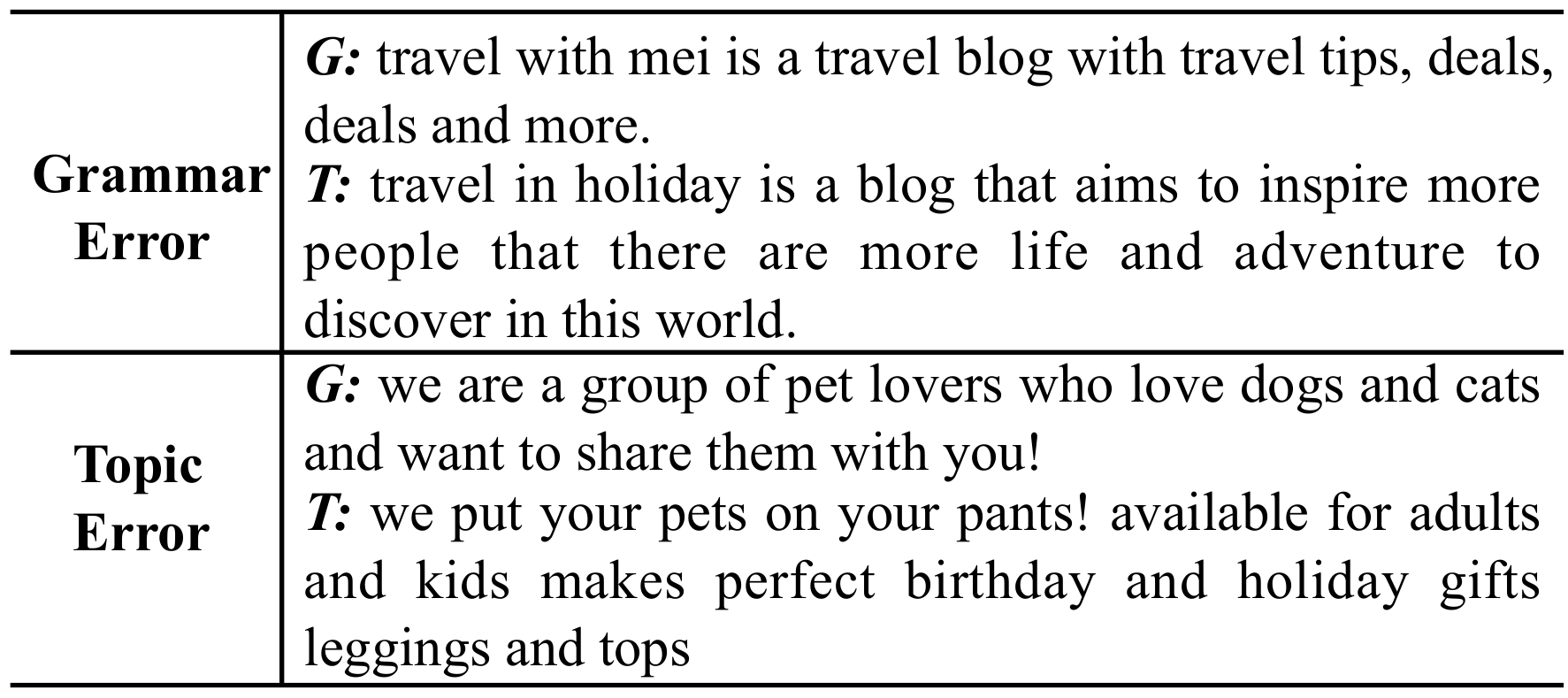}
\vspace{-1.75em}
\caption{Examples of major error types for the generation results of UTGED (G) and the target reference (T).
}
 \vspace{-0.5em}
\label{fig:error}
\end{figure}

%% file: Sections/conclusion.tex
\section{Conclusion}
We have presented a new application to generate personalized self-introduction, where a large-scale Twitter dataset is gathered for experiments.
A novel unified topic-guided encoder-decoder framework is proposed to leverage latent topics for distilling essential user interests from the numerous noisy tweets a user has posted.
Empirical results show our model outperforms advanced PLM-based models, shedding light on the potential of latent topics in helping PLMs digest lengthy and noisy input.

%% file: Sections/ack.tex
\section*{Acknowledgements}
This paper is substantially supported by the NSFC Young Scientists Fund (No.62006203, 62106105), a grant from the Research Grants Council of the Hong Kong Special Administrative Region, China (Project No. PolyU/25200821), and the Innovation and Technology Fund  (Project No. PRP/047/22FX).

%% file: Sections/limitation.tex
\section*{Limitations}
First, inference efficiency is one of the main limitations of this work. 
The BART model takes about 14 minutes to complete the inference on our dataset, while our UTGED needs 92 minutes. 
The reason for the slow inference is that UTGED requires heavy computation to update the gradient to the encoder's states and decoder's states (as shown in Eq.\ref{eq:atribute_loss}\textasciitilde Eq.\ref{eq:updated_state}). 
Future work may consider how to advance model efficiency further.

Second, the lack of multimodal content in the published tweets would result in another limitation. The images contained in the published tweets are ignored in this work. However, due to the complicated relationships between images and texts in a multimodal tweet, images might provide complementary content and complete the meanings of the message \cite{DBLP:conf/acl/VempalaP19}. Therefore, future studies might explore self-introduction generation using multimodal tweets (images and text) to indicate personal interests.

%% file: Sections/Ethics_Statement.tex
\section*{Ethics Statement}
Our paper constructs a large-scale Twitter dataset for a self-introduction generation. The data acquisition procedure follows the standard data collection process regularized by Twitter API. 
Only the public users and tweets are gathered.
The downloaded data is only used for academic research. For our experiments, the data has been anonymized for user privacy protection, e.g., authors’ names are removed, @mention and URL links are changed to common tags. Following Twitter’s policy for content redistribution, we will only release the anonymized data. Additionally, we will require data requestors to sign a declaration form before obtaining
the data, ensuring that the dataset will only be
reused for the purpose of research, complying with Twitter’s data policy, and not for gathering anything that probably raises ethical issues, such as sensitive personal information. For the human annotations, we recruited the annotators as part-time research assistants with 15 USD/hour payment.

%% file: Sections/appendix.tex
\section{Appendix}
\label{sec:appendix}
\subsection{Tweet Selection Algorithm} \label{alg:selection_algorithm}
\input{figure_sections/pseudocode_filter.tex}

\subsection{Data filtering}
\label{append:data_filtering}
\input{Table/removed_data_score.tex}
Here we show the original dataset's distribution before filtering based on the similarity score: [-1, 0): 9,680; [0, 0.1): 123,560; [0.1, 0.2): 257,759; [0.2, 0.3): 193,847; [0.3, 0.4): 157,478; [0.4, 0.5): 118,881; [0.5, 0.6): 44,455; [0.6, 0.7): 10,977; [0.7, 0.8): 1,691; [0.8, 0.9): 169; [0.9, 1.0]: 26. We observe that the number of user samples first increases and then decreases, indicating that self-introductions are related to the user's historical tweets. Otherwise, the data distribution will tend to exhibit a long tail (based on social media characteristics). 

Additionally, we tested a sample of 10,000 users with similarity scores fall in the ranges of [0.3,0.4), [0.2,0.3), [0.1, 0.2), [0,0.1) and results of the best model $\text{Oracle}_{A}$+Topic on low-similarity data samples are shown in Table \ref{tab:low_similarity_score}. The results indicate that low-similarity data samples do impact negatively on the training results.

\subsection{Full Experimental Results} \label{append:full_results}
\paragraph{Varying Topic Number.}
We show the results from BART+S+E on the left of ``/'' and those from UTGED on the right.
\input{Table/nc_results.tex}

\paragraph{Varying Prompt length.}
We show the results from BART+S+E on the left of ``/'' and those from UTGED on the right.
\input{Table/prompt_len.tex}

\paragraph{Varying Sentence Number.}
We show the results from BART+S on the left of ``/'' and those from UTGED on the right.
\input{Table/sentence_num.tex}
\newpage
\subsection{Topic Words} \label{append:topic2words}
\input{figure_sections/topic2words.tex}

\newpage
\subsection{Detailed Case Study} \label{append:longer_case}
\input{figure_sections/case_study_full.tex}

%% file: figure_sections/pseudocode_filter.tex
\begin{algorithm}[h]
  \caption{Selecting representative tweets}
  \label{alg:sent_filter}
  \begin{algorithmic}[1]
    \Require
      collected tweets pool for user $u^i$: $X^i=\{x^i_1,x^i_2,...,x^i_m\}$
    \Ensure
      representative tweets shortlist $R^i$
    \State initial $R^i=\{\}$
    \Repeat
        \State calculate overall similarity score between 
        \Statex $\qquad$ $\qquad$ a tweet and other tweets in $X^i$;
        \State assume the tweet with the highest score is \Statex $\qquad$ $\qquad$ $x^i_h$, remove $x^i_h$ from $X^i$ to $R^i$;
        \State calculate the similarity score between $x^i_h$ \Statex $\qquad$ $\qquad$ and remained tweets in $X^i$;
        \State for the tweets whose similarity score is \Statex $\qquad$ $\qquad$ higher than $\lambda$, remove it from $X^i$;
    \Until{there are no tweets in $X^i$}
  \end{algorithmic}
\end{algorithm}

%% file: Table/removed_data_score.tex
\begin{table}[h]
	 \centering
{\renewcommand{\arraystretch}{1.0}
\resizebox{0.8\columnwidth}{!}
{
	\begin{tabular}[b]{l c c c}
		\toprule
		\textbf{Similarity Score}        & \textbf{R-1} & \textbf{R-2} &\textbf{R-L}  \\
		\midrule
		
     	$[0,0.1)$ &  8.43  &  0.91  &  7.67  \\
            $[0.1,0.2)$ & 11.49  & 1.94  & 10.50  \\
            $[0.2,0.3)$ & 17.87   & 4.50   & 15.03  \\
            $[0.3,0.4)$ & 24.25  & 7.52 & 21.35   \\
		\bottomrule	
  \end{tabular}}}
	\caption{
	The results of $\text{Oracle}_{A}$+Topic on low-similarity data samples. 
	}
	\label{tab:low_similarity_score}
\end{table}

%% file: Table/nc_results.tex
\begin{table}[h]
	 \centering
{\renewcommand{\arraystretch}{1.0}
\resizebox{1.0\columnwidth}{!}
{
	\begin{tabular}[b]{l c c c}
		\toprule
		\textbf{K}        & \textbf{R-1} & \textbf{R-2} &\textbf{R-L}  \\
		\midrule
		
     	50 & 24.58/24.85  &  7.85/7.99  &  21.49/21.76  \\
      100 & 24.78/24.99  & 7.95/8.05  & 21.65/21.84  \\
            150 & 24.77/24.98   & 7.93/8.01   & 21.60/21.78  \\
            200 & 24.67/24.71  & 7.86/7.90  & 21.52/21.59   \\

		\bottomrule	\end{tabular}}}
	\caption{
	The effects of topic number $K$. 
	}
	\label{tab:topic_num_results}
\end{table}

%% file: Table/prompt_len.tex
\begin{table}[h]
	 \centering
{\renewcommand{\arraystretch}{1.0}
\resizebox{1.0\columnwidth}{!}
{
	\begin{tabular}[b]{l c c c}
		\toprule
		\bm{$L$}        & \textbf{R-1} & \textbf{R-2} &\textbf{R-L}  \\
		\midrule
		
            3 & 24.36/24.53  & 7.68/7.78  & 21.25/21.42  \\
            7 & 24.78/24.99  & 7.95/8.05  & 21.65/21.84   \\
            11 & 24.48/24.69  & 7.90/8.01  & 21.43/21.66  \\
            15 & 24.53/24.74  & 7.84/7.89  & 21.41/21.58   \\
            19 & 24.34/24.48  &7.76/7.81   & 21.30/21.42    \\
	    
		\bottomrule	\end{tabular}}}
	\caption{
	The effects of prompt length $L$. 
	}
	\label{tab:prompt_len_results}
\end{table}

%% file: Table/sentence_num.tex
\begin{table}[h]
	 \centering
{\renewcommand{\arraystretch}{1.0}
\resizebox{1.0\columnwidth}{!}
{
	\begin{tabular}[b]{l c c c}
		\toprule
		\textbf{SN}        & \textbf{R-1} & \textbf{R-2} &\textbf{R-L}  \\
		\midrule
		
     	20 & 22.79/23.41  &  6.90/7.15  &  19.92/20.49  \\
      40 & 23.82/24.57  & 7.45/7.83  & 20.81/21.50  \\
            60 & 24.12/24.90   & 7.63/8.02   & 21.09/21.78  \\
            80 & 24.23/24.95  & 7.70/8.06  & 21.15/21.82   \\
            100 & 24.26/24.99  & 7.68/8.05  & 21.17/21.84  \\
	    
		\bottomrule	\end{tabular}}}
	\caption{
	The effects of sentence number (SN). 
	}
	\label{tab:sent_num_results}
\end{table}

%% file: figure_sections/topic2words.tex
\begin{figure}[h]
\centering
\includegraphics[width=1.0\columnwidth]{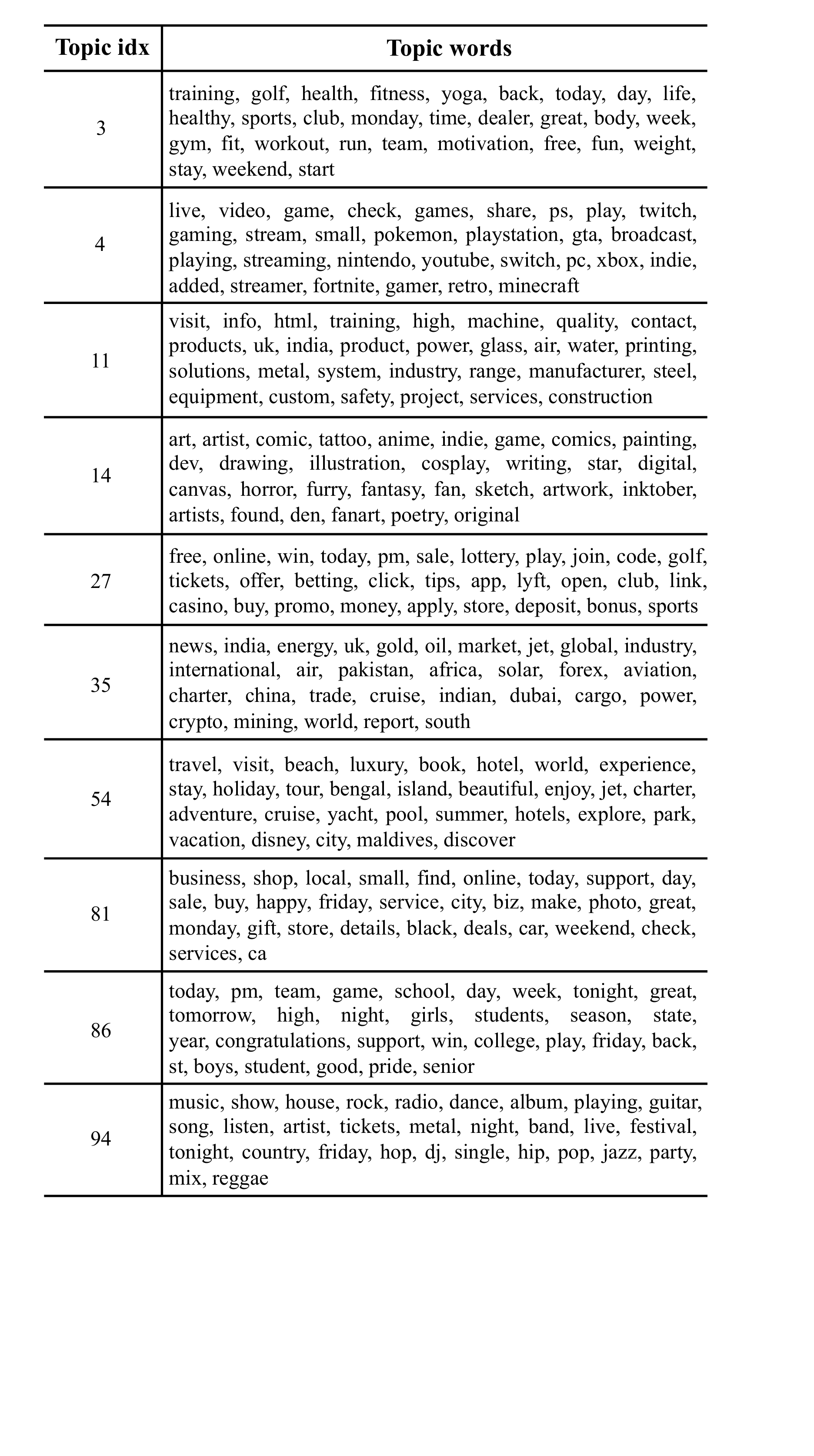}
\caption{Randomly sampled 10 topics with their top-30 topic words.
}
 \vspace{-1.5em}
\label{fig:topic2words}
\end{figure}

%% file: figure_sections/case_study_full.tex
\begin{figure}[h]
\centering
\includegraphics[width=1\columnwidth]{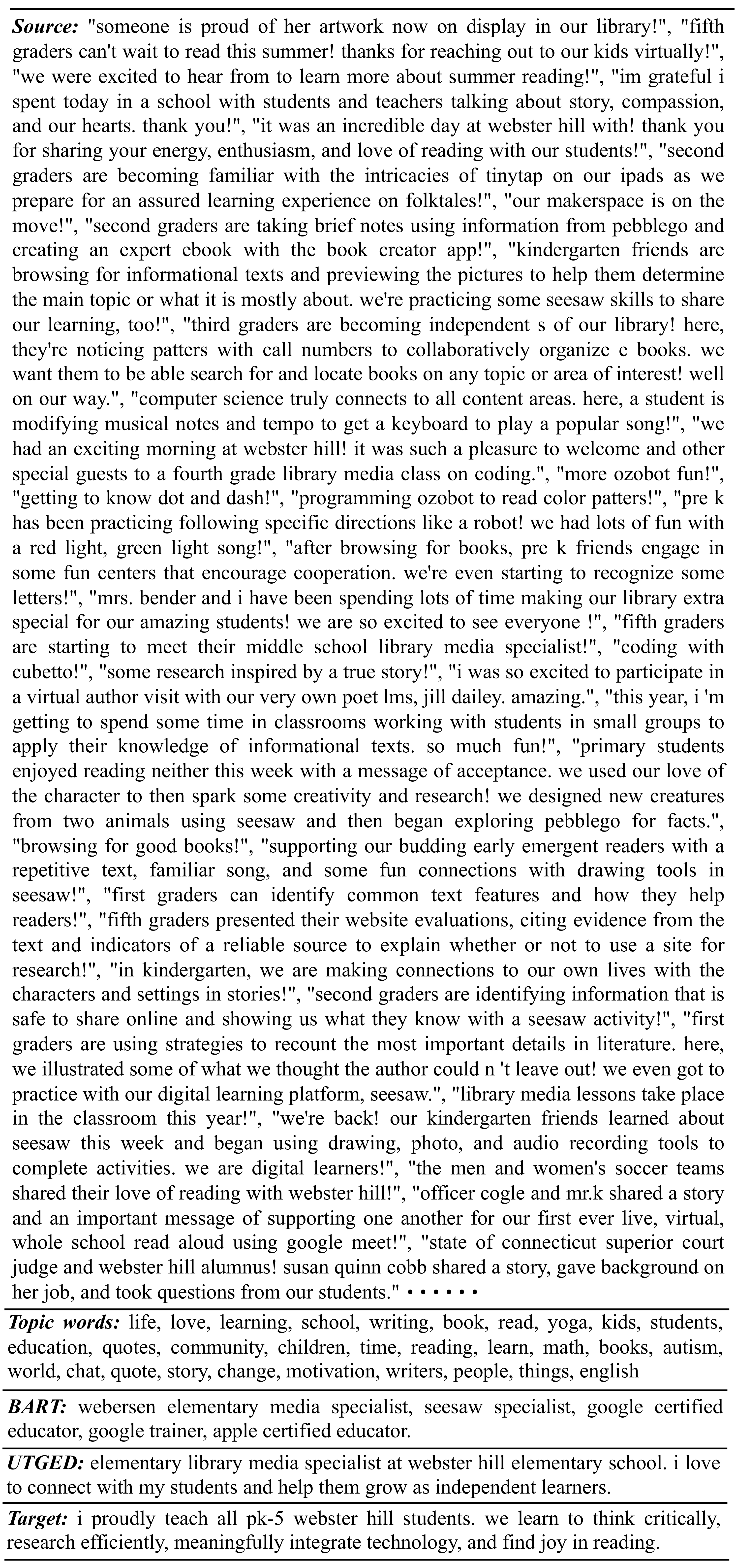}
\vspace{-2em}
\caption{
A Twitter user sample and the related results. 
From top to down shows user history (source $T^i$), topic words ($A^i$), BART output, UTGED output, and reference self-introduction (target $Y^i$). The source text consists of 70 tweets, and here we randomly sample half of them to put in the figure for a better display.
}
 \vspace{-1.5em}
\label{fig:case_study_full}
\end{figure}